\def\year{2018}\relax
\pdfoutput=1
\documentclass[letterpaper]{article}
\usepackage{aaai18}
\usepackage{times}
\usepackage{helvet}
\usepackage{courier}
\usepackage{url}
\usepackage{graphicx}
\usepackage{amsfonts,amsmath,amssymb,mathtools}
\usepackage[export]{adjustbox}
\DeclareMathOperator*{\argmin}{arg\,min}
\newcommand{\inR}[2]{\in \mathbb{R}^{#1 \times #2}}
\newcommand{\citet}[1]{\citeauthor{#1} ̃\shortcite{#1}}
\newcommand{\citep}{\cite}

\begin{document}

\title{Improving the Adversarial Robustness and Interpretability of Deep Neural
       Networks by Regularizing their Input Gradients}

\author{
  Andrew Slavin Ross \and Finale Doshi-Velez\\
  Paulson School of Engineering and Applied Sciences, Harvard University, Cambridge, MA 02138, USA\\
  andrew\_ross@g.harvard.edu, finale@seas.harvard.edu}

\maketitle

\begin{abstract}
  Deep neural networks have proven remarkably effective at solving many
  classification problems, but have been criticized recently for two major
  weaknesses: the reasons behind their predictions are uninterpretable, and the
  predictions themselves can often be fooled by small adversarial perturbations. These
  problems pose major obstacles for the adoption of neural networks in domains
  that require security or transparency.
  In this work, we evaluate the effectiveness of defenses that
  differentiably penalize the degree to which small changes in inputs can alter model
  predictions. Across multiple attacks,
  architectures, defenses, and datasets, we find that neural networks trained
  with this \emph{input gradient} regularization exhibit robustness to transferred
  adversarial examples generated to fool all of the other models. We also find
  that adversarial examples generated to fool gradient-regularized models fool
  all other models equally well, and actually lead to more ``legitimate,''
  interpretable misclassifications as rated by people (which we confirm in a
  human subject experiment). Finally, we demonstrate that regularizing input gradients
  makes them more naturally interpretable as rationales for
  model predictions.  We conclude by discussing this relationship
  between interpretability and robustness in deep neural networks.
\end{abstract}

\section{Introduction}

Over the past several years, progress in training deep neural networks (DNNs)
has greatly expanded the scope of what machine learning models can accomplish.
However, especially as they start to be used in settings which are
security-sensitive or have legal ramifications
\cite{kang2017prediction}, many in the field have noted important
problems that fall into two major categories.

The first is that DNNs can be easily manipulated
into making incorrect predictions on carefully doctored examples which, to
humans, look indistinguishable from examples it classifies correctly
\cite{szegedy2013intriguing}. Although many techniques
for generating these examples (which we call ``attacks'') require access to
model parameters, \citet{papernot2017practical} have shown that it is possible
and even practical to attack black-box models in the real world, in large part
because of the \textit{transferability} of adversarial examples; examples
generated to fool one model tend to fool \textit{all} models trained on the
same dataset. Particularly for images, these adversarial examples can be
constructed to fool models across a variety of scales and perspectives
\cite{athalye2017synthesizing}, which poses a problem for the adoption of deep
learning models in systems like self-driving cars.

Although there has recently been a great deal of research in adversarial
defenses, many of these methods have struggled to achieve robustness to
transferred adversarial examples \cite{tramer2017space}. Some of the most effective defenses,
such as feature squeezing \cite{xu2017feature}, simply detect and reject adversarial examples
rather than making predictions.
The most
common, ``brute force'' solution is adversarial training, where we simply include a mixture of
normal and adversarially-generated examples in the training set
\cite{kurakin2016adversarial}.  However, \citet{tramer2017ensemble} show that
the robustness adversarial training provides can be circumvented by randomizing or transferring
perturbations from other models (though ensembling helps).

In addition to concerns about robustness, domain experts are also often concerned that DNN predictions are uninterpretable.  The lack of interpretability is particularly problematic in
domains where algorithmic bias is often a factor \cite{propub} or in
medical contexts where safety risks can arise when there is mismatch between how a model is trained and used
\cite{caruana2015intelligible}. Cases
like these have motivated research in \textit{explaining} DNN predictions,
which can reveal their implicit biases \cite{adler2016auditing} or alert a
domain expert that a prediction was made for the wrong reasons. The form these
explanations often take is an interpretable local surrogate model, often a
linear model, which simulates how the network will respond to small
perturbations of its inputs \cite{lime}.

One choice for generating these local linear models is simply to take the
model's gradient with respect to its inputs, which provides a local
linear approximation of the model's behavior \cite{baehrens2010explain}. However, especially
for image classification tasks, few researchers examine the raw input gradients
directly because they are noisy and difficult to interpret.  This issue has
spurred the development of techniques like integrated gradients
\cite{integrated-gradients} and SmoothGrad
\cite{smoothgrad} that generate smoother, more interpretable saliency maps
from noisy gradients. The rationale behind these techniques is that, while the
local behavior of the model may be noisy, examining the gradients over larger
length scales in input space provides a better intution about the model's
behavior.

However, raw input gradients are \textit{exactly} what many attacks use to
generate adversarial examples. Explanation techniques which smooth out
gradients in background pixels may be inappropriately hiding the fact that the
model is quite sensitive to them.  We consider that perhaps the need for these smoothing
techniques in the first place is indicative of a problem with our models, related to their adversarial vulnerability and capacity to overfit. Perhaps it is fundamentally hard for adversarially vulnerable models to be fully interpretable.

On the other hand, perhaps it is hard for interpretable models to be adversarially vulnerable.
Our hypothesis is that by training a model to have smooth input
gradients with fewer extreme values, it will not only be more interpretable but
also more resistant to adversarial examples. In the experiments that
follow we confirm this hypothesis using gradient regularization,
which directly optimizes the model to have smooth
input gradients with respect to its predictions during training.
Using gradient regularization, we demonstrate robustness to adversarial
examples across multiple model architectures and datasets, and in particular
demonstrate robustness to \textit{transferred} adversarial examples: gradient-regularized models maintain significantly higher accuracy on examples
generated to fool other models than baselines. Furthermore, both
qualitatively and in human subject experiments, we find that
adversarial examples generated to fool gradient-regularized models are, in a
particular sense, more ``interpretable'': they fool humans as well.

\section{Background}

In this section, we will introduce notation, and give a brief overview of the
baseline attacks and defenses against which we will test and
compare our methods. The methods we will analyze apply to all differentiable
classification models $f_\theta(X)$, which are functions parameterized by
$\theta$ that return predictions $\hat{y} \inR{N}{K}$ given inputs $X
\inR{N}{D}$. These predictions indicate the probabilities that each of $N$
inputs in $D$ dimensions belong to each of $K$ class labels. To train these
models, we try to find sets of parameters $\theta^*$ that minimizs the total
information distance between the predictions $\hat{y}$ and the true labels $y$
(also $\inR{N}{K}$, one-hot encoded) on a training set: \begin{equation}
\begin{split}
  \theta^* & = \argmin_\theta \sum_{n=1}^N \sum_{k=1}^K -y_{nk} \log f_\theta(X_n)_k, \\
\end{split}
\label{eq:normal-training}
\end{equation}
which we will sometimes write as \[
  \argmin_\theta H(y, \hat{y}),
\]
with $H$ giving the sum of the cross entropies between the predictions and the
labels.

\subsection{Attacks}

\subsubsection{Fast Gradient Sign Method (FGSM)}

\citet{fgsm} introduced this first method of generating adversarial examples by
perturbing inputs in a manner that increases the local linear approximation of
the loss function:
\begin{equation}
X_{\text{FGSM}} = X + \epsilon\, \text{sign}\left(\nabla_x H(y, \hat{y})\right)
\label{eq:fgsm}
\end{equation}
If $\epsilon$ is small, these adversarial examples are indistinguishable from
normal examples to a human, but the network performs significantly worse on
them.

\citet{tgsm} noted that one can iteratively perform this attack with a small
$\epsilon$ to induce misclassifications with a smaller total perturbation (by
following the nonlinear loss function in a series of small linear steps rather
than one large linear step).

\subsubsection{Targeted Gradient Sign Method (TGSM)}

A simple modification of the Fast Gradient Sign Method is the Targeted Gradient
Sign Method, introduced by \citet{tgsm}. In this attack, we attempt to decrease
a modified version of the loss function that encourages the model to
misclassify examples in a specific way:
\begin{equation}
  X_{\text{TGSM}} = X - \epsilon\, \text{sign}\left(\nabla_x H(y_{\text{target}}, \hat{y})\right),
\label{eq:tgsm}
\end{equation}
where $y_{\text{target}}$ encodes an alternate set of labels we would like the
model to predict instead. In the digit classification experiments below, we
often picked targets by incrementing the labels $y$ by 1 (modulo 10), which we
will refer to as $y_{+1}$. The TGSM can also be performed iteratively.

\subsubsection{Jacobian-based Saliency Map Approach (JSMA)}

The final attack we consider, the Jacobian-based Saliency Map Approach (JSMA), also
takes an adversarial target vector $y_{\text{target}}$. It iteratively searches
for pixels or pairs of pixels in $X$ to change such that the probability of
the target label is increased and the probability of all other labels are
decreased. This method is notable for producing examples that have only been
changed in several dimensions, which can be hard for humans to detect. For a
full description of the attack, we refer the reader to \citet{jsma}.

\subsection{Defenses}

As baseline defenses, we consider defensive distillation and adversarial training.
To simplify comparison, we omit defenses \cite{xu2017feature,nayebi2017biologically} that are not fully
architecture-agnostic or which work by detecting and rejecting adversarial examples.

\subsubsection{Distillation}

Distillation, originally introduced by \citet{ba2014distillation},
was first examined as a potential
defense by \citet{distillation}. The main idea is that we train the model twice, initially using the one-hot ground truth labels but ultimately using the initial model's softmax probability
outputs, which contain additional information about the problem. Since the normal softmax
function tends to converge very quickly to one-hot-ness, we divide all of the logit network outputs
(which we will call $\hat{z}_k$ instead of the
probabilities $\hat{y}_k$) by a temperature $T$ (during training but not evaluation): \begin{equation}
  f_{T,\theta}(X_n)_k = \frac{e^{\hat{z}_k(X_n)/T}}{\sum_{i=1}^K e^{\hat{z}_i(X_n)/T}},
\end{equation}
where we use $f_{T,\theta}$ to denote a network ending in a softmax with temperature $T$.
Note that as $T$ approaches $\infty$, the predictions converge to
$\frac{1}{K}$. The full process can be expressed as \begin{equation}
\begin{split}
  \theta^0 & = \argmin_\theta \sum_{n=1}^N \sum_{k=1}^K -y_{nk} \log f_{T,\theta}(X_n)_k, \\
  \theta^* & = \argmin_\theta \sum_{n=1}^N \sum_{k=1}^K -f_{T,\theta^0}(X_n)_k \log f_{T,\theta}(X_n)_k.
\end{split}
\label{eq:distillation}
\end{equation}
Distillation is usually used to help small networks achieve the same accuracy as larger DNNs,
but in a defensive context, we use the same model twice.
It has been shown to be an effective defense against white-box FGSM attacks,
but \citet{carlini2016defensive} have shown that
it is not robust to all kinds of attacks. We will see that the precise way it
defends against certain attacks is qualitatively different than gradient regularization, and that it can actually make the models more vulnerable to attacks than an undefended model.

\subsubsection{Adversarial Training}

In adversarial training \cite{kurakin2016adversarial}, we increase robustness
by injecting adversarial examples into the training procedure. We follow the method implemented in
\citet{papernot2016cleverhans},
where we augment the network to run the FGSM on the training batches and
compute the model's loss function as the average of its loss on normal and
adversarial examples without allowing gradients to propogate so as to weaken the FGSM attack (which would also make the method second-order). We compute FGSM perturbations with respect to
predicted rather than true labels to prevent ``label leaking,'' where our model
learns to classify adversarial examples more accurately than regular examples.

\section{Input Gradient Regularization}

Input gradient regularization is a very old idea. It was first introduced by \citet{doublebackprop}
as ``double backpropagation'', which trains neural networks by minimizing not just the ``energy'' of the network but the
rate of change of that energy with respect to the input features.
In their formulation the energy is a
quadratic loss, but we can formulate it almost equivalently using the
cross-entropy: \begin{equation}
\begin{split}
  \theta^* & = \argmin_\theta \sum_{n=1}^N \sum_{k=1}^K -y_{nk} \log f_\theta(X_n)_k \\
                                   & +  \lambda \sum_{d=1}^D \sum_{n=1}^N \left(\frac{\partial}{\partial x_d} \sum_{k=1}^K -y_{nk} \log f_\theta(X_n)_k\right)^2,
\end{split}
\label{eq:doubleback}
\end{equation}
whose objective we can write a bit more concisely as \[
  \argmin_\theta H(y, \hat{y}) + \lambda ||\nabla_x H(y, \hat{y})||_2^2,
\]
where $\lambda$ is a hyperparameter specifying the penalty strength.
The goal of this approach is to ensure that if any input changes slightly, the
KL divergence between the predictions and the labels will not change significantly. Double
backpropagation was mentioned as a potential adversarial defense in the same
paper which introduced defensive distillation \cite{distillation}, but to our
knowledge, its effectiveness in this respect has not yet been analyzed in the
literature.

Note that it is also possible to regularize the input gradients of different
cross entropies besides $H(y, \hat{y})$. For example, if we replace $y$ with a
uniform distribution of $\frac{1}{K}$ for all classes (which we will abbreviate
as $H(\frac{1}{K}, \hat{y})$), then we penalize the sensitivity of the
divergence between the predictions and uniform uncertainty; we will call this
penalty ``certainty sensitivity.''  Certainty sensitivity can also be interpreted as the score
function of the predictions with respect to the inputs.
Certainty sensitivity penalties have been
used to stabilize the training of Wasserstein GANs \cite{wgans} and to
incorporate domain knowledge-specific regularization \cite{rrr}.
We explore the relative performance of different gradient regularization techniques
at different $\lambda$ in Figure \ref{fig:method-comp}.

\section{Experiments}

\subsubsection{Datasets and Models}

We evaluated the robustness of distillation, adversarial training, and
gradient regularization to the FGSM, TGSM, and JSMA on MNIST \cite{mnist}, Street-View House Numbers (SVHN)
\cite{svhn}, and notMNIST \cite{not-mnist}.  On all datasets, we test a simple
convolutional neural network with 5x5x32 and 5x5x64 convolutional layers
followed by 2x2 max pooling and a 1024-unit fully connected layer, with
batch-normalization after all convolutions and both batch-normalization and
dropout on the fully-connected layer. All models were implemented in
Tensorflow and trained using Adam \cite{adam} with $\alpha=0.0002$ and $\epsilon=10^{-4}$ for 15000 minibatches of size of 256.
For SVHN, we prepare training and validation set as described in
\citet{sermanet2012convolutional}, converting the images to grayscale following
\citet{grundland2007decolorize} and applying both global and local contrast
normalization.

\subsubsection{Attacks and Defenses}

\begin{figure}[ht]
  \centering
  \includegraphics[width=0.49\textwidth]{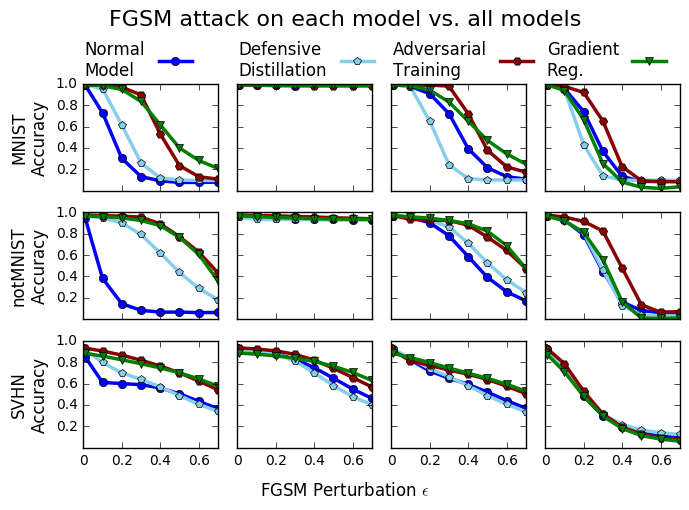}
  \caption{Accuracy of all CNNs on FGSM examples generated to fool undefended
    models, defensively distilled, adversarially trained, and
    gradient regularized models (from left to right) on MNIST, SVHN, and
    notMNIST (from top to bottom). Gradient-regularized models are the most
    resistant to other models' adversarial examples at high $\epsilon$, while all models are fooled
    by gradient-regularized model examples. On MNIST and notMNIST, distilled model examples are usually identical to non-adversarial examples (due to gradient underflow), so they fail to fool any of the other models.
  }
\label{fig:fgsm-comp}
\end{figure}

\begin{figure}[ht]
  \centering
  \includegraphics[width=0.49\textwidth]{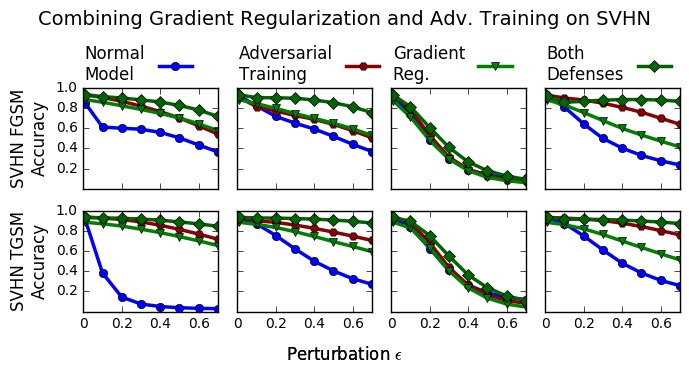}
  \caption{Applying both gradient regularization and adversarial training (``both defenses'') allows us to obtain maximal robustness to white-box and normal black-box attacks on SVHN (with a very slight label-leaking effect on the FGSM, perhaps due to the inclusion of the $\nabla_x H(y,\hat{y})$ term). However, no models are able to maintain robustness to black-box attacks using gradient regularization.}
\label{fig:both-defenses}
\end{figure}

For adversarial training and JSMA example generation, we used the Cleverhans adversarial example library
\cite{papernot2016cleverhans}. For distillation, we used a softmax temperature
of $T=50$, and for adversarial training, we trained with FGSM perturbations at
$\epsilon=0.3$, averaging normal and adversarial losses.
For gradient regularized models, we use double backpropagation, which provided the best robustness, and train over a spread of $\lambda$ values. We choose the $\lambda$ with the highest accuracy against validation black-box FGSM
examples but which is still at least 97\% as accurate on normal validation
examples (though accuracy on normal examples tended not to be significantly different). We explore the effects of varying $\lambda$ in Figure \ref{fig:traincurves-by-lambda}. Code for all models and experiments has been open-sourced \footnote{https://github.com/dtak/adversarial-robustness-public}.

\subsubsection{Evaluation Metrics}

For the FGSM and TGSM, we test \textit{all} models
against adversarial examples generated for \textit{each} model and report accuracy.
Testing this way allows us to simultaneously measure white- and black-box robustness.

On the JSMA and iterated TGSM, we found that measuring accuracy was no longer
a good evaluation metric, since for our gradient-regularized models, the
generated adversarial examples often resembled their targets more than their
original labels. To investigate this, we performed a human subject experiment
to evaluate the legitimacy of adversarial example misclassifications.

\subsection{Accuracy Evaluations (FGSM and TGSM)}

\subsubsection{FGSM Robustness}

Figure \ref{fig:fgsm-comp} shows the results of our defenses' robustness to the
FGSM on MNIST, SVHN, and notMNIST for our CNN at a variety of perturbation strengths
$\epsilon$.
Consistently across datasets, we find that gradient-regularized models exhibit
strong robustness to transferred FGSM attacks (examples produced by attacking other models). Although adversarial training sometimes performs slightly better at
$\epsilon \leq 0.3$, the value we used in training, gradient regularization generally surpasses it at higher $\epsilon$.

Interestingly, although gradient-regularized models seem vulnerable to white-box attacks,
they actually fool \textit{all} other models equally well.
In this respect, gradient regularization may hold promise not just as a defense but as an
attack, if examples generated to fool them are inherently more transferable.

Models trained with defensive distillation in general perform no better and
often worse than undefended models. Remarkably, except on SVHN, attacks against
distilled models actually fail to fool all models. Closer inspection of distilled model gradients and
 examples themselves reveals that this occurs because distilled FGSM
gradients vanish -- so the examples are not perturbed at all. As soon as we obtain a nonzero perturbation
from a different model, distillation's appearance of robustness vanishes as well.

Although adversarial training and gradient regularization seem comparable in terms of accuracy,
they work for different reasons and can be applied in concert to increase robustness, which we show in Figure \ref{fig:both-defenses}. In Figure \ref{fig:overlaps} we also show that, on normal and adversarially trained black-box FGSM attacks, models trained with these two defenses are fooled by different sets of adversarial examples.

\begin{figure}[htb]
  \centering
  \includegraphics[width=0.49\textwidth]{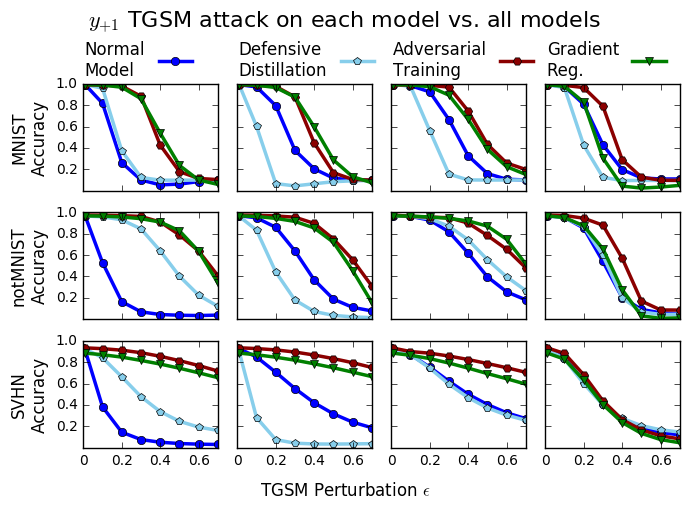}
  \caption{CNN accuracy on $y_{+1}$ TGSM examples generated to fool the four
  models on three datasets (see Figure \ref{fig:fgsm-comp} for more
explanation). Gradient-regularized models again exhibit robustness to
other models' adversarial examples. Distilled model adversarial perturbations fool other models again since their input gradients no longer underflow.}
\label{fig:tgsm-comp}
\end{figure}

\subsubsection{TGSM Robustness}

Against the TGSM attack (Figure \ref{fig:tgsm-comp}), defensively distilled model gradients no
longer vanish, and
accordingly these models start to show the same vulnerability to adversarial attacks as
others.  Gradient-regularized models still exhibit the same
robustness even at large perturbations $\epsilon$, and again, examples
generated to fool them fool other models equally well.

\begin{figure}[htb]
  \centering
  \includegraphics[width=0.5\textwidth]{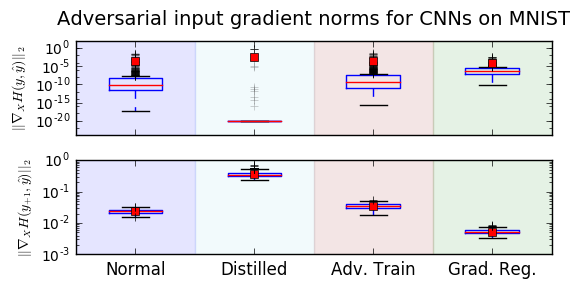}
  \includegraphics[width=0.5\textwidth]{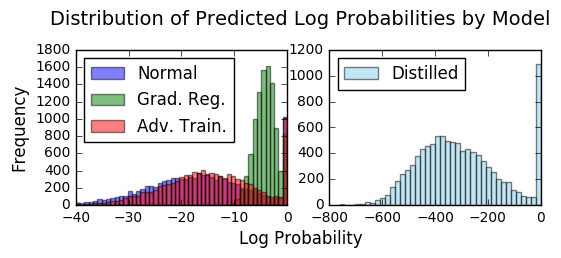}
  \caption{Distributions of (L2 norm) magnitudes of FGSM input gradients (top), TGSM input gradients (middle),
    and predicted log probabilities across all classes (bottom) for each defense. Note the logarithmic scales. Gradient-regularized models
    tend to assign non-predicted classes higher probabilities, and the L2 norms of the input gradients of their FGSM and TGSM
  loss function terms have similar orders of magnitude. Distilled models (evaluated at $T=0$) assign extremely small probabilities to all but the predicted class, and their TGSM gradients explode while their FGSM gradients vanish (we set a minimum value of $10^{-20}$ to prevent underflow). Normal and adversarially trained models lie somewhere in the middle.}
\label{fig:mnist-gradient-comparison}
\end{figure}

One way to better understand the differences between gradient-regularized, normal,
and distilled models is to examine the log probabilities they output and the norms of their loss function input gradients, whose
distributions we show in Figure \ref{fig:mnist-gradient-comparison} for MNIST. We can see that the different defenses have very different statistics.
Probabilities of non-predicted classes tend to be small but remain nonzero for gradient-regularized models,
while they vanish on defensively distilled models evaluated at $T=0$ (despite distillation's stated purpose of discouraging certainty). Perhaps because $\nabla \log p(x) = \frac{1}{p(x)}\nabla p(x)$, defensively distilled models' non-predicted log probability input gradients are the largest by many orders of magnitude, while gradient-regularized models' remain controlled, with much smaller means and variances (see Figure \ref{fig:traincurves-by-lambda} for a visualization of how this behavior changes with $\lambda$). The other models lie between these two extremes.
While we do not have a strong theoretical argument about what input gradient magnitudes
\textit{should} be, we believe it makes intuitive sense that having less
variable, well-behaved, and non-vanishing input gradients should be associated
with robustness to attacks that consist of small perturbations in input space.

\begin{figure}[htb]
  \centering
  \includegraphics[width=\linewidth]{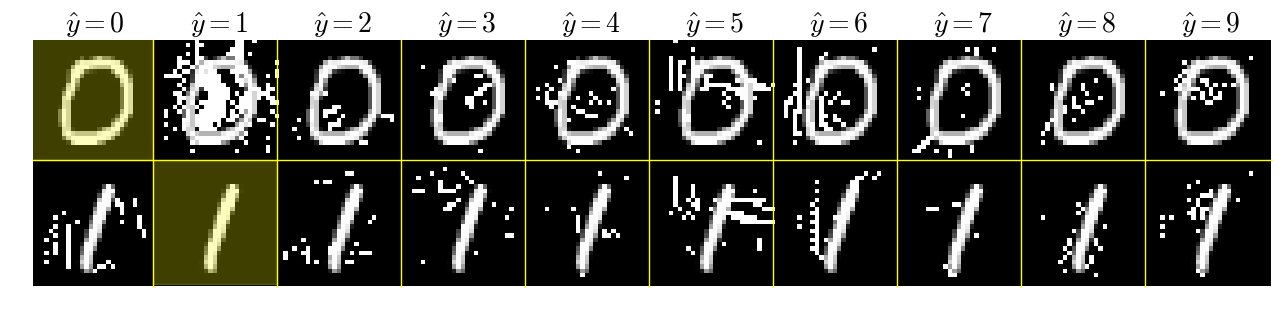}\\
  \includegraphics[width=\linewidth]{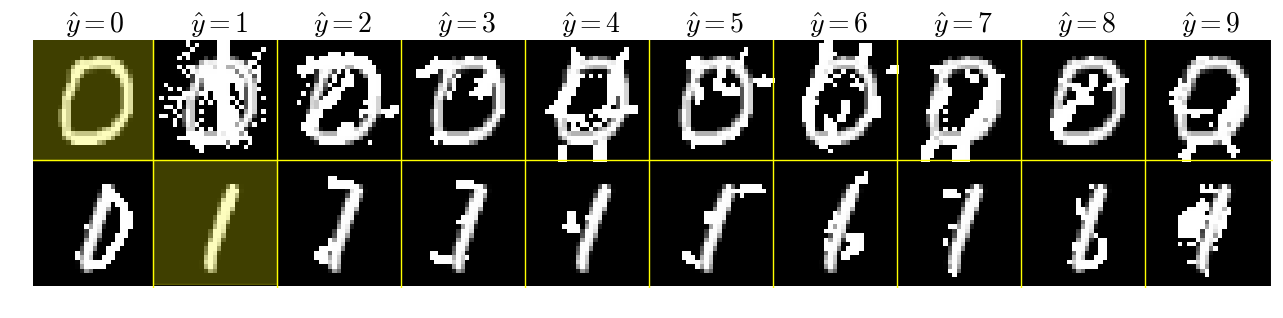}
  \caption{Results of applying the JSMA to MNIST \texttt{0} and \texttt{1}
  images with maximum distortion parameter $\gamma=0.25$ for a distilled model
(top) and a gradient-regularized model (bottom). Examples in each row start
out as the highlighted digit but are modified until the model predicts the
digit corresponding to their column or the maximum distortion is reached.}
  \label{fig:mnist-jsma-grid}
\end{figure}

\subsection{Human Subject Study (JSMA and Iterated TGSM)}

\subsubsection{Need for a Study}

Reporting accuracy numbers for the JSMA can be misleading, since without a
maximum distortion constraint it necessarily runs until the model predicts the
target.  Even with such a constraint, the perturbations it creates sometimes
alter the examples so much that they no longer resemble their original labels,
and in some cases bear a greater resemblance to their targets.  Figure
\ref{fig:mnist-jsma-grid} shows JSMA examples on MNIST for
gradient-regularized and distilled models which attempt to convert \texttt{0}s
and \texttt{1}s into every other digit.  Although all of the perturbations
``succeed'' in changing the model's prediction, we can see that in the
gradient-regularized case, many of the JSMA examples strongly resemble their
targets.

The same issues occur for other attack methods, particularly the iterated TGSM,
for which we show confusion matrices for different models and datasets in the Appendix in Figure \ref{fig:tgsm-grids}.
For the gradient-regularized models, these psuedo-adversarial examples often
represent intuitive transformations of one digit into another, which is not
reflected in accuracies with respect to the original labels.

To test these intuitions more rigorously, we ran a small pilot study with 11
subjects to measure whether they found examples generated by these methods to
be more or less plausible instances of their targets.

\subsubsection{Study Protocol}

The pilot study consisted of a quantitative and qualitative portion. In the
quantitative portion, subjects were shown 30 images of MNIST JSMA or SVHN
iterated TGSM examples. Each of the 30 images corresponded to one original
digit (from 0 to 9) and one model (distilled, gradient-regularized, or
undefended). Note that for this experiment, we used $\nabla_x H(\frac{1}{K}, \hat{y})$ gradient regularization and trained models for 4 epochs at a learning rate of 0.001, which was sufficient to produce examples with explanations similar to the longer training procedure used in our earlier experiments, and actually increased the robustness of the undefended models (adversarial accuracy tends to fall with training iteration).
Images were chosen uniformly at random from a
larger set of 45 examples that corresponded to the first 5 images of the
original digit in the test set transformed using the JSMA or iterated TGSM to
each of the other 9 digits (we ensured that all models misclassified all
examples as their target). Subjects were not given the original label, but were
asked to input what they considered the most and second-most plausible
predictions for the image that they thought a reasonable classifier would make
(entering N/A if they thought no label was a plausible choice).  In the
qualitative portion that came afterwards, users were shown three 10x10
confusion matrices for the different defenses on MNIST (Figure
\ref{fig:mnist-jsma-grid} shows the first two rows) and were asked to write
comments about the differences between the examples.  Afterwards, there was a
short group discussion. This study was performed in compliance with the
institution's IRB.

\begin{table}
  \begin{tabular}{|c|p{0.35in}|p{0.55in}|p{0.35in}|p{0.55in}|} \hline
 & \multicolumn{2}{c|}{MNIST (JSMA)} & \multicolumn{2}{|c|}{SVHN (TGSM)} \\
\hline
Model & human\newline fooled &   mistake\newline reasonable   & human\newline fooled & mistake\newline reasonable  \\
\hline
normal      & 2.0\%           & 26.0\%           & 40.0\%          & 63.3\%         \\
\hline
distilled   & 0.0\%           & 23.5\%           & 1.7\%           & 25.4\%          \\
\hline
grad. reg. & \textbf{16.4\%} & \textbf{41.8\%}  & \textbf{46.3\%} & \textbf{81.5\%}  \\
\hline
\end{tabular}
\caption{Quantitative feedback from the human subject experiment. ``human fooled'' columns record what percentage of examples were classified by humans as \textit{most} plausibly their adversarial targets, and ``mistake reasonable'' records how often humans either rated the target plausible or marked the image unrecognizable as any label (N/A).}
\label{table:human-study}
\end{table}

\subsubsection{Study Results}

Table \ref{table:human-study} shows quantitative results from the human subject
experiment. Overall, subjects found gradient-regularized model adversarial examples most
convincing. On SVHN and especially MNIST, humans were most likely to think that
gradient-regularized (rather than distilled or normal) adversarial examples were
best classified as their target rather than their original digit.
Additionally, when they did not consider the target the most plausible
label, they were most likely to consider gradient-regularized model mispredictions
``reasonable'' (which we define in Table \ref{table:human-study}), and more
likely to consider distilled model mispredictions unreasonable.
p-values for the differences between normal and gradient regularized unreasonable error rates were 0.07 for MNIST and 0.08 for SVHN.

In the qualitative portion of the study (comparing MNIST JSMA examples), \textit{all} of the written
responses described significant differences between the insensitive model's
JSMA examples and those of the other two methods. Many of the examples for the
gradient-regularized model were described as ``actually fairly convincing,'' and that the
normal and distilled models ``seem to be most easily fooled by adding spurious
noise.'' Few commentators indicated any differences between the normal and
distilled examples, with several saying that ``there doesn't seem to be [a]
stark difference'' or that they ``couldn't describe the difference'' between
them. In the group discussion one subject remarked on how the perturbations to
the gradient-regularized model felt ``more intentional'', and others commented on how
certain transitions between digits led to very plausible fakes while others
seemed inherently harder. Although the study was small, both its quantitative and qualitative
results support the claim that gradient regularization, at least for the two
CNNs on MNIST and SVHN, is a credible defense against the JSMA and the iterated
TGSM, and that distillation is not.

\subsection{Connections to Interpretability}

\begin{figure*}[htb]
  \centering
  \parbox{0.32\linewidth} {
    \includegraphics[width=\linewidth]{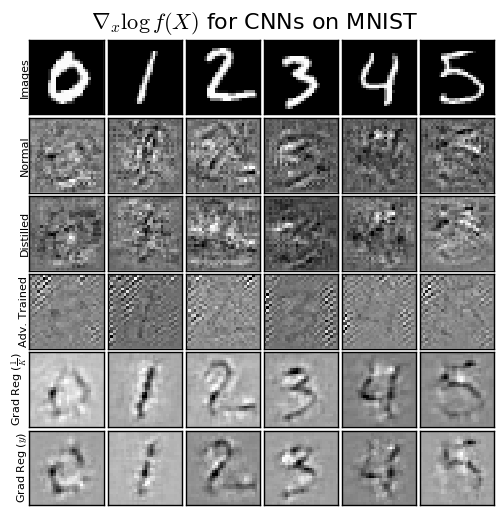}
  }
  \parbox{0.32\linewidth} {
    \includegraphics[width=\linewidth]{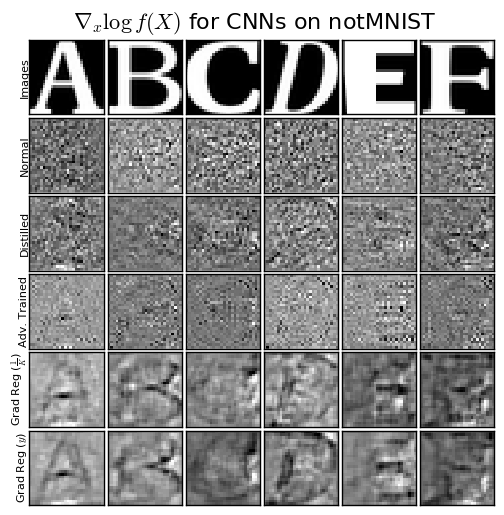}
  }
  \parbox{0.32\linewidth} {
    \includegraphics[width=\linewidth]{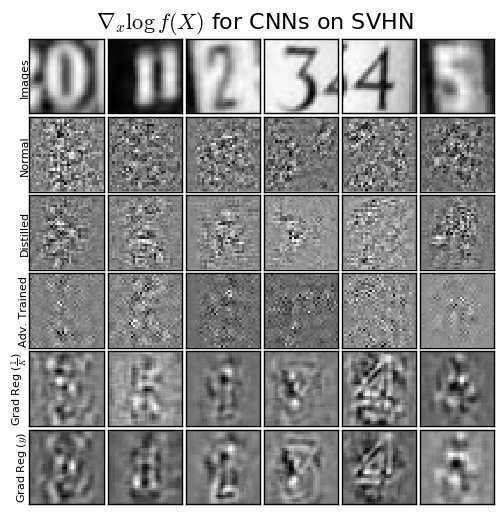}
  }
  \caption{Input gradients of the model's score function $\nabla_x H(\frac{1}{K},\hat{y})$
    that provide a local linear approximation of normal
    models (top), distilled models at $T=50$ (second from top), adversarially trained models (middle),
    and models trained with $\nabla_x H(\frac{1}{K},\hat{y})$ and $\nabla_x H(y,\hat{y})$
    gradient regularization (bottom two). Whitening black pixels or darkening
    white pixels makes the model more certain of its prediction. In general, regularized model
    gradients appear smoother and make more intuitive sense as local linear
    approximations.}
  \label{fig:uncertainty-gradients}
\end{figure*}

Finally, we present a qualitative evaluation suggesting a connection between
adversarial robustness and interpretability.  In the literature on
explanations, input gradients are frequently used as explanations
\cite{baehrens2010explain}, but sometimes they are noisy and not interpretable
on their own. In those cases, smoothing techniques have been developed
\cite{smoothgrad,integrated-gradients} to generate more interpretable
explanations, but we have already argued that these techniques may obscure
information about the model's sensitivity to background features.

We hypothesized that if the models had more interpretable input gradients without the need for
smoothing, then perhaps their adversarial examples, which are generated
directly from their input gradients, would be more interpretable as well.  That is, the adversarial example would be more obviously transformative away from the original class label and towards
another. The results of the user study show that our gradient-regularized
models have this property; here we ask if the gradients more interpretable as
explanations.

In Figure \ref{fig:uncertainty-gradients} we visualize input
gradients across models and datasets, and while we cannot
make any quantitative claims, there does appear to be a qualitative
difference in the interpretability of the input gradients between the
gradient-regularized models (which were
relatively robust to adversarial examples) and the normal and distilled models
(which were vulnerable to them). Adversarially trained models seem to exhibit slightly more interpretable gradients,
but not nearly to the same degree as gradient-regularized models.
When we repeatedly apply input gradient-based perturbations using the iterated TGSM (Figure \ref{fig:tgsm-grids}), this difference in interpretability between models is greatly magnified, and the results for gradient-regularized models seem to provide insight into what the model has learned. When gradients are interpretable, we may actually be able to use adversarial attacks as explanations.

\section{Discussion}

In this paper, we have analyzed the performance of gradient regularization,
a novel way of training differentiable models that penalizes the degree to which
infinitesimal changes to $X$ affect predictions $\hat{y}$. We have shown
that training with gradient regularization increases
robustness to adversarial perturbations as much or more than adversarial training,
and that the two methods can be combined to achieve even greater robustness.
We also showed that input gradient regularization increases
the interpretability of adversarial perturbations as rated by
human subjects. Although a larger study that also tests
adversarial training would be necessary to verify this,
our results strongly suggest that the way distillation and adversarial training
increase robustness is qualitatively different than gradient regularization,
and not associated with greater interpretability.

There is ample opportunity to improve gradient regularization. Although we
explored performance for several variants in Figure \ref{fig:method-comp},
there are many alternative formulations we could explore, including gradient
penalties on logits rather than log probabilities, or a wider variety of
example-specific cross-entropies.  It may also be the case that network
hyperparameters should be different when
training networks with gradient regularization. Future work should explore
this avenue, as well as testing on larger, more state-of-the-art networks.

One weakness of gradient regularization is that it is a second-order
method; including input gradients in parameter gradient descent requires taking
second derivatives, which in our experiments increased training time per batch
by a factor of slightly more than 2. Figures \ref{fig:traincurves-by-lambda} and \ref{fig:traincurves-by-regmethod} also suggest they may take longer to converge.
Distillation, of course, requires twice as
much training time by definition, and adversarial training increased train time by closer to a factor of 3
(since we had to evaluate, differentiate, and re-evaluate each batch). However,
input gradient regularization increases the size of the computational graph in a way that
the other methods do not, and second derivatives are not always supported for all operations
in all autodifferentiation frameworks. Overall, we feel that the
increase in training time for gradient regularization is manageable,
but it still comes at a cost.

What we find most promising about gradient regularization, though,
is that it significantly changes the shape of
the models' decision boundaries, which suggests that they make predictions for
qualitatively different (and perhaps better) reasons. It is unlikely that
regularizing for this kind of smoothness will be a panacea for all
manifestations of the ``Clever Hans'' effect \cite{horse} in deep neural
networks, but in this case the prior it represents -- that
predictions should not be sensitive to small perturbations in input space
-- helps us find models that make more robust and interpretable predictions. No
matter what method proves most effective in the general case, we suspect that
any progress towards ensuring either interpretability or adversarial robustness
in deep neural networks will likely represent progress towards both.

\section*{Acknowledgements}

We thank Nicolas Papernot and Weiwei Pan for helpful discussions and
comments. We also thank our anonymous reviewers for their comments
which helped us gain insights into the generality of gradient-based
regularization.  FDV acknowledges support from AFOSR FA9550-17-1-0155
and we acknowledge the Berkman Klein Center.

\bibliographystyle{aaai}
\bibliography{bibliography}

\begin{appendix}

\section*{Appendix}

\begin{figure*}
  \centering
  \includegraphics[width=\textwidth]{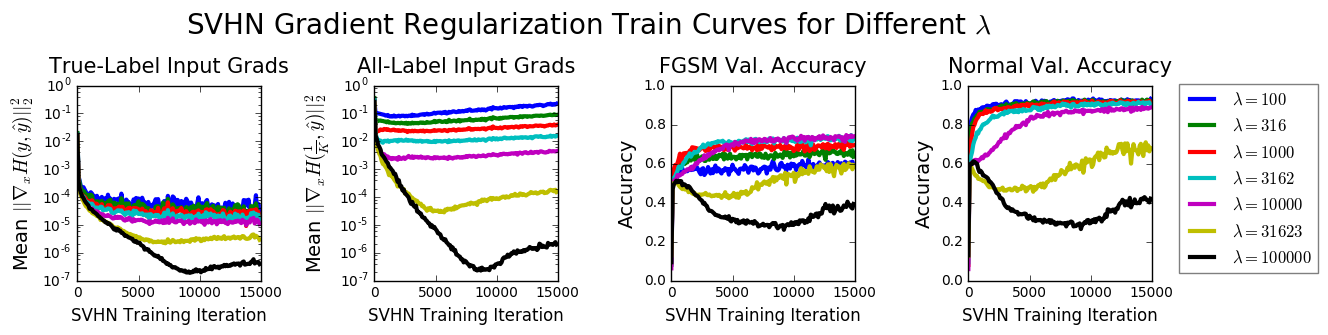}
  \caption{Variation in several important training quantities by gradient regularization penalty strength $\lambda$. In general, less regularized models tend to have larger input gradients (left two plots) and lower FGSM validation accuracy (second from right), while more regularized models have smaller input gradients and better FGSM accuracy. However, they take longer to reach high normal accuracy (right), and it is possible to over-regularize them (bottom two curves). Over-regularized models tend to have equal gradients with respect to all log probabilities (as well as equal normal and adversarial accuracy).}
  \label{fig:traincurves-by-lambda}
\end{figure*}

\begin{figure*}
  \centering
  \includegraphics[width=\textwidth]{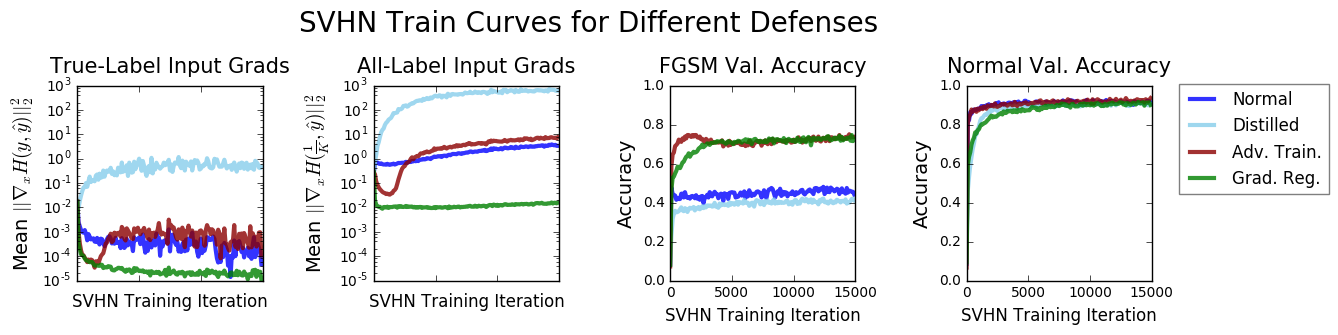}
  \caption{Variation in training quantities across defense method. Although distillation in general causes true-label input gradients $\nabla_x H(y,\hat{y})$ to vanish for most examples, enough other examples' gradients explode to ensure the average value remains highest. Adversarially trained model input gradients are similar in magnitude to normal model input gradients, suggesting they are more similar to normal than gradient-regularized models.}
  \label{fig:traincurves-by-regmethod}
\end{figure*}

\begin{figure*}
  \centering
  \includegraphics[width=0.48\textwidth,valign=t]{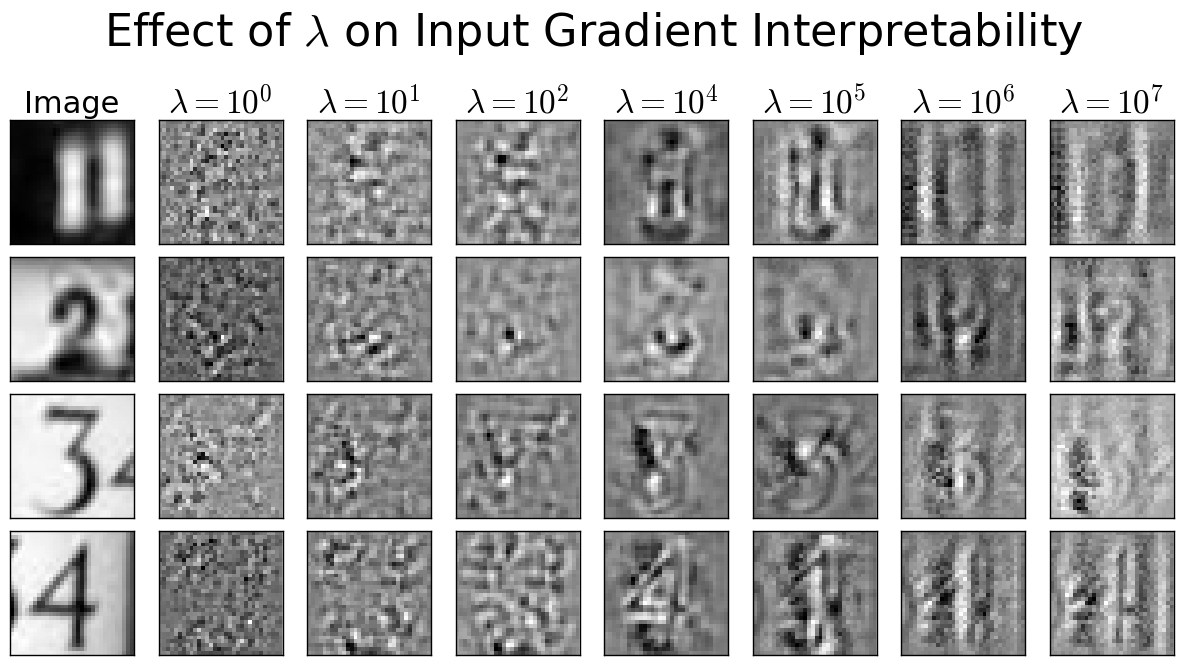}
  \,
  \includegraphics[width=0.48\textwidth,valign=t]{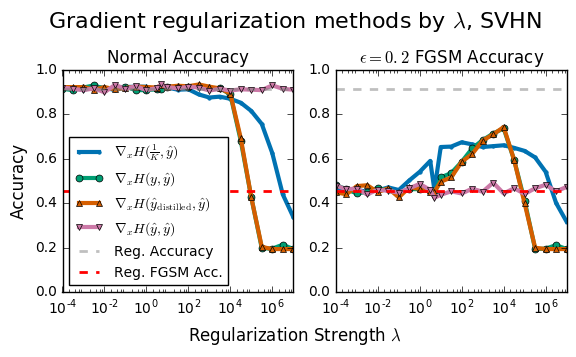}

  \caption{
    Left: $\nabla_x H(\frac{1}{K},\hat{y})$ input gradients of a double backpropagation-trained model on SVHN examples as we vary $\lambda$. Gradient interpretability tends to track FGSM accuracy.
    Right: Accuracies of different gradient regularization methods as we vary $\lambda$. In general, we find that double backpropagation (penalizing predicted log probabilities $\nabla_x H(y,\hat{y})$) obtains slightly better FGSM accuracy than certainty insensitivity (penalizing the sum of log probabilities $\nabla_x H(\frac{1}{K}, \hat{y})$), though it was sometimes more sensitive to the value of $\lambda$. We also noticed the certainty sensitivity penalty sometimes destabilized our training procedure (see the blip at $\lambda=10^1$). To see if we could combine distillation and gradient regularization, we also tried using distilled probabilities $\hat{y}_{\rm{distilled}}$ obtained from a normal CNN trained and evaluated at $T=50$, which was comparable to but no beterr than double backpropagation. Using the model's own probability outputs $\hat{y}$ (i.e. its entropy) seemed to have no effect.
}
  \label{fig:method-comp}
\end{figure*}

\begin{figure*}
  \centering
  \includegraphics[width=\textwidth]{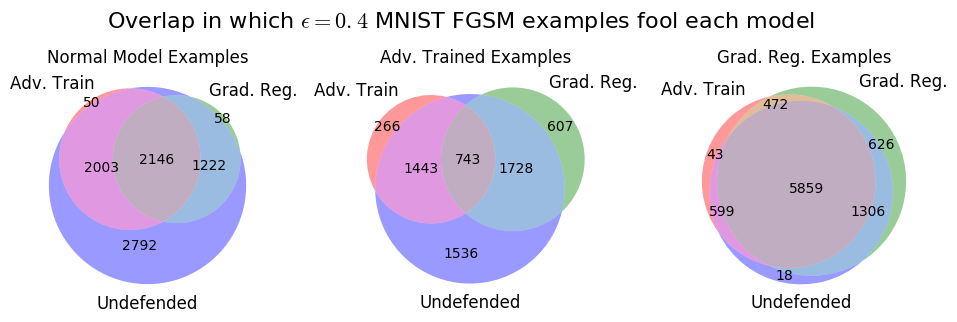}
  \caption{Venn diagrams showing overlap in which MNIST $\epsilon=0.4$ FGSM examples, generated for normal, adversarially trained, and gradient regularized models, fool all three. Undefended models tend to be fooled by examples from all models, while the sets of adversarially trained model FGSM examples that fool the two defended models are closer to disjoint. Gradient-regularized model FGSM examples fool all models. These results suggest that ensembling different forms of defense may be effective in defending against black box attacks (unless those black box attacks use a gradient-regularized proxy).}
  \label{fig:overlaps}
\end{figure*}

\begin{figure*}
  \centering
  \includegraphics[width=0.32\textwidth]{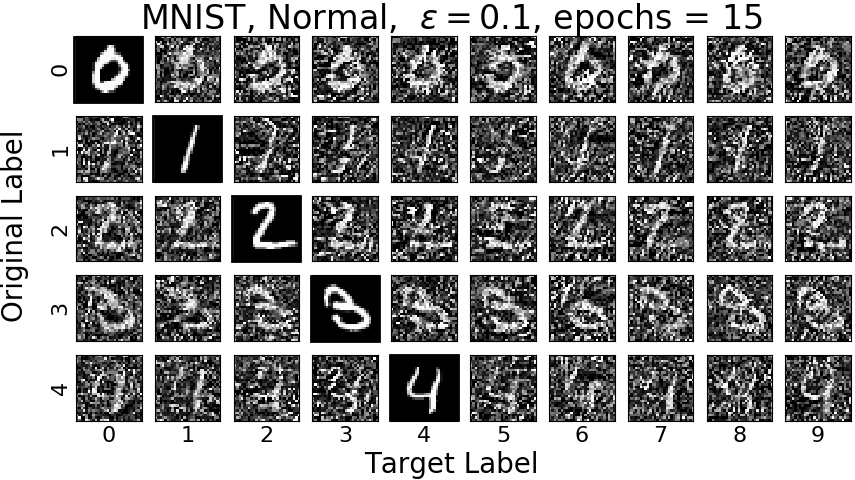}
  \includegraphics[width=0.32\textwidth]{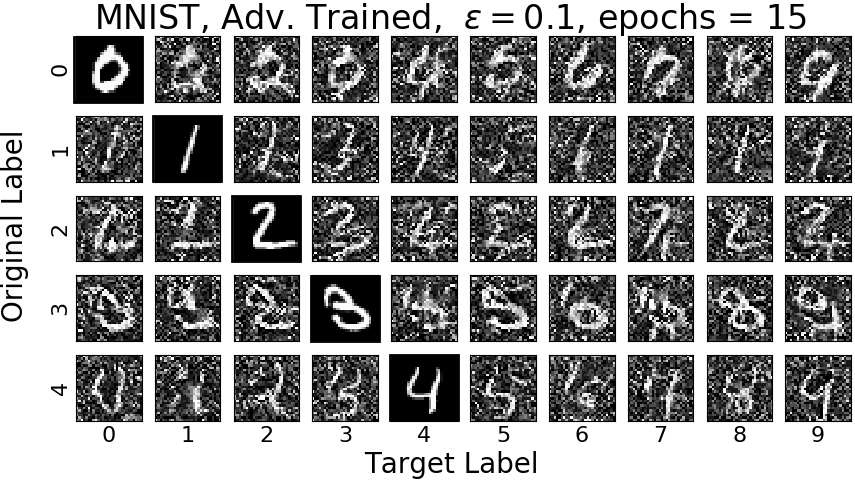}
  \includegraphics[width=0.32\textwidth]{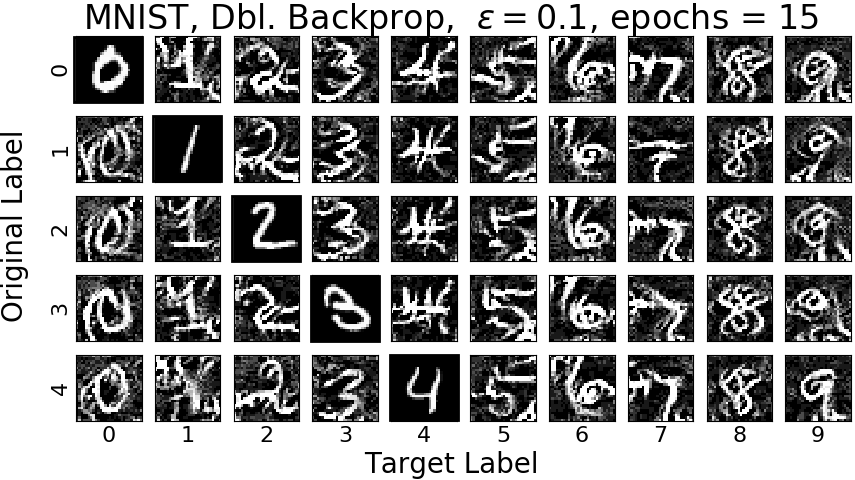}

  \includegraphics[width=0.32\textwidth]{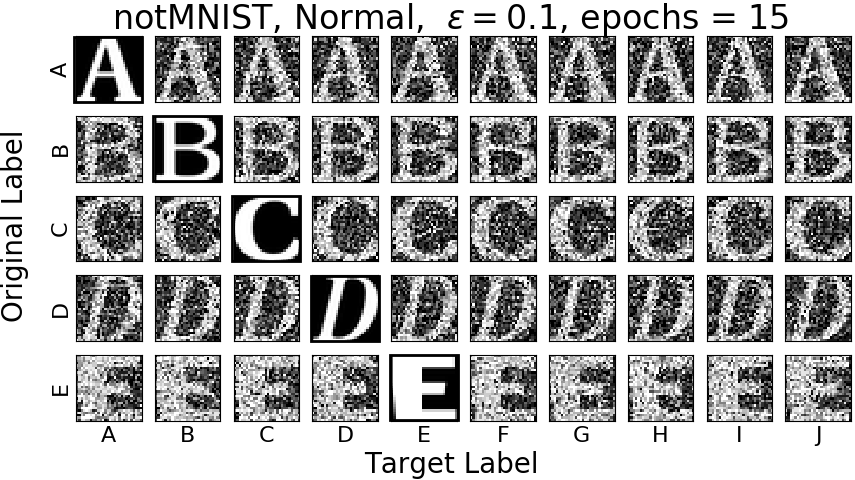}
  \includegraphics[width=0.32\textwidth]{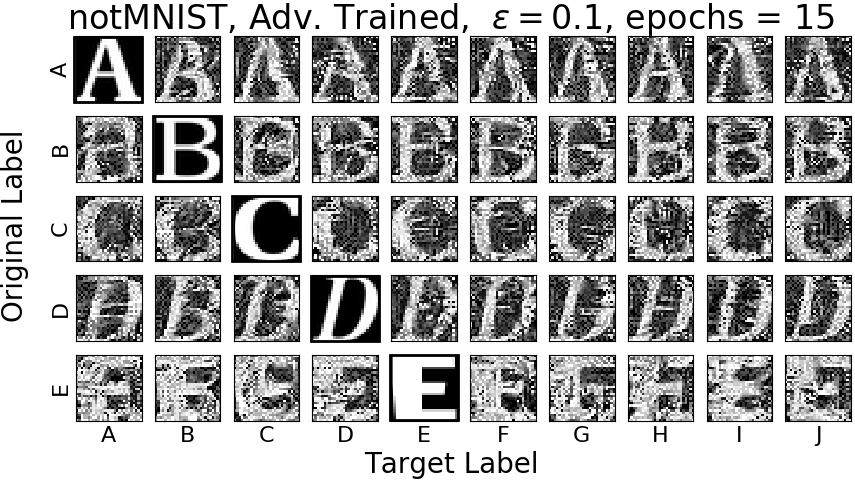}
  \includegraphics[width=0.32\textwidth]{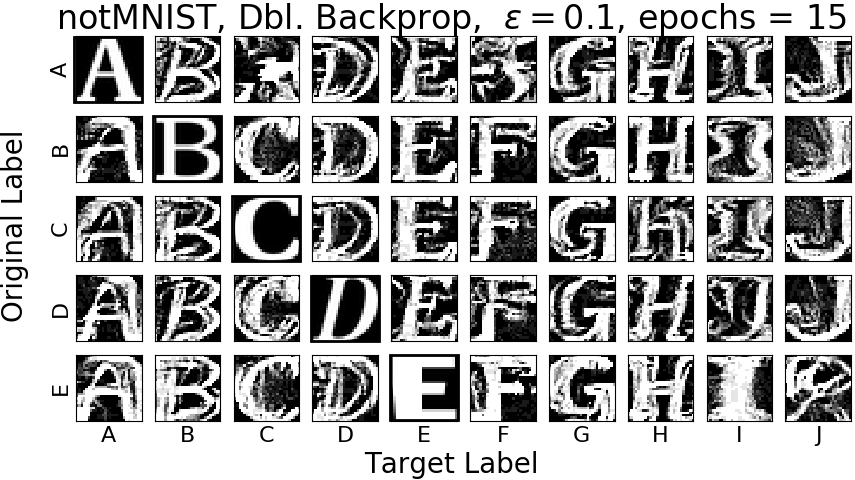}

  \includegraphics[width=0.32\textwidth]{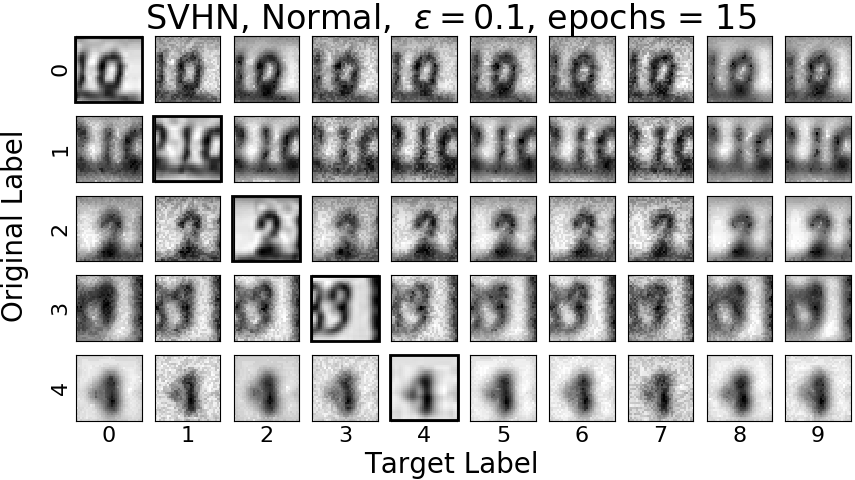}
  \includegraphics[width=0.32\textwidth]{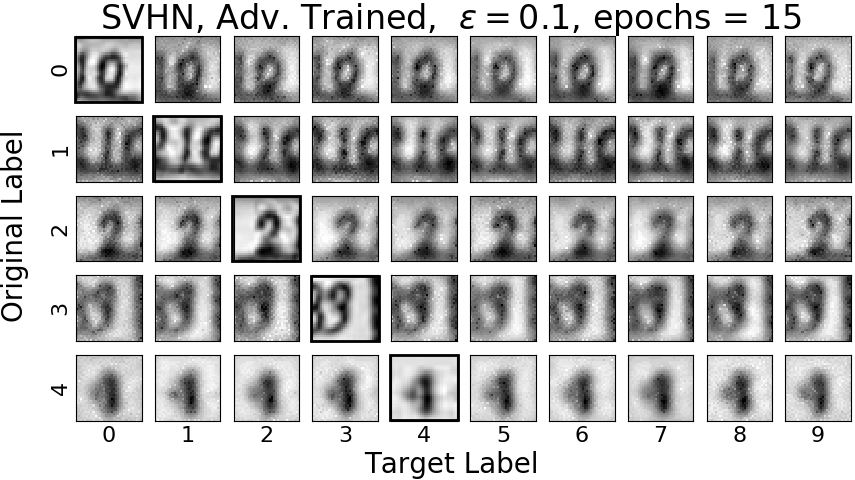}
  \includegraphics[width=0.32\textwidth]{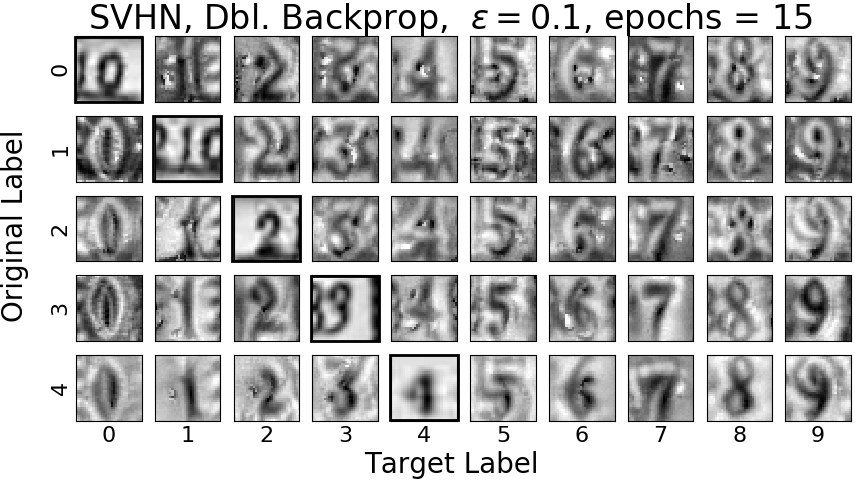}

  \caption{
    Partial confusion matrices showing results of applying the iterated TGSM for 15 iterations at $\epsilon=0.1$. Each row is generated from the same example but modified to make the model to predict every other class. TGSM examples generated for gradient-regularized models (right) resemble their targets more than their original labels and may provide insight into what the model has learned. Animated versions of these examples can be seen at \texttt{http://goo.gl/q8ZM1T}, and code to generate them is available at \texttt{http://github.com/dtak/adversarial-robustness-public}.
}
  \label{fig:tgsm-grids}
\end{figure*}

\end{appendix}
\end{document}